\documentclass{article}
\usepackage{ijcai26}

\usepackage{times}
\usepackage{soul}
\usepackage{url}
\usepackage[hidelinks]{hyperref}
\usepackage[utf8]{inputenc}
\usepackage[small]{caption}
\usepackage{graphicx}
\usepackage{amsmath}
\usepackage{amsthm}
\usepackage{booktabs}
\usepackage{algorithm}
\usepackage{algorithmic}
\usepackage[switch]{lineno}

\title{Evaluating Federated Pre-Training: On the Reliability of Downstream Fine-Tuning and Intrinsic Evaluation}

\author{
Claudia Großer$^{\textbf{*}1,2}$
\and
Maike Heuer$^{\textbf{*}1,2}$\and
Denis Krompaß$^{2}$\And
Thomas A. Runkler$^{1,2}$\\
\affiliations
$^1$Technical University Munich\\
$^2$Siemens AG\\
\emails
\{claudia.grosser, maike.heuer, denis.krompass, thomas.runkler\}@siemens.com
}

\begin{document}

\maketitle

\begin{abstract}
    Federated pre-training offers a way to train foundation models on private or distributed data without centralizing the underlying datasets.
    However, evaluating federated pre-training remains challenging because differences in client participation and local data availability can make directly comparable evaluation difficult. Moreover, pre-training test perplexity is tied to the pre-training distribution, while downstream benchmarks introduce task-specific adaptation that may not faithfully reflect the test perplexity established during pre-training.
    In this work, we study which evaluation protocol more reliably reflects federated pre-training quality. Using a controlled set of centralized and federated-trained models of a 16M parameter transformer model trained on identical client data, we assess evaluation protocols by whether they preserve a reference ranking established on the same pre-training testset. We compare downstream fine-tuning on GLUE, including full, head-only, and reduced-data variants, with next-token prediction on GLUE text as an intrinsic evaluation signal. Our results show that downstream fine-tuning does not reliably preserve the pre-training ranking, whereas direct next-token prediction exhibits a strong correspondence with the pre-training test perplexity. These findings suggest that downstream fine-tuning alone can be misleading when comparing federated pre-trained models, and that evaluation signals closer to the original pre-training objective deserve greater attention.
\end{abstract}

\section{Introduction}
Foundation models are at the core of modern AI, such as GPT \cite{singh2025openai} or Claude \cite{anthropic2025system}. These models are typically obtained through large-scale pre-training on vast amounts of data. Yet many valuable data sources, especially in domains such as healthcare or industry, cannot easily be incorporated into centralized pre-training pipelines because they are privacy-sensitive, regulated, or institutionally siloed.

Federated learning \cite{FedAvg} offers a natural alternative in such settings. Instead of collecting data centrally, model training is performed across distributed clients while the underlying data remain local. This makes federated learning a promising approach for pre-training foundation models on data that would otherwise remain inaccessible to centralized training. Consequently, a crucial question arises: \textbf{How can the quality of federated pre-training be effectively evaluated?}

Evaluating pre-trained models presents a distinct set of considerations compared to assessing task-specific systems. If the goal is to assess the effectiveness of the pre-training procedure itself, an evaluation protocol must isolate and highlight differences directly attributable to the pre-training phase, rather than those introduced during subsequent supervised adaptation. 
This is particularly relevant for federated foundation models, as robust evaluation methods are essential for advancing novel federated pre-training strategies, incentivizing clients and for enabling informed model selection and deployment. Researchers and practitioners need clear ways to compare and improve various federated pre-training algorithms and configurations, ensuring that new approaches yield superior models when working with decentralized data. Furthermore, an effective evaluation framework is essential for making informed decisions about which pre-trained model is best when pre-training has occurred on different training corpora.

While evaluating pre-trained models involves downstream fine-tuning on benchmark tasks, this approach, though useful for measuring end-task performance, may not reliably reflect the true quality of the pre-training. The fine-tuning process can inadvertently conflate the effects of the initial pre-training with those of task-specific adjustments, potentially obscuring a clear comparison of pre-training effectiveness.

In this work, we evaluate a controlled set of centralized and federated 16M parameter transformer models trained on identical client data. Using the aggregated client test sets, we establish a reference ranking after pre-training and then assess whether alternative evaluation protocols preserve this ordering and expected relative differences between pre-trained models.
Although our study is conducted on smaller 16M parameter transformer models in a controlled setting, it is intended as a first step toward identifying an evaluation issue that should be considered in the study of federated pre-training of foundation models more broadly.

\begin{figure}
  \centering
  \includegraphics[width=1.0\linewidth]{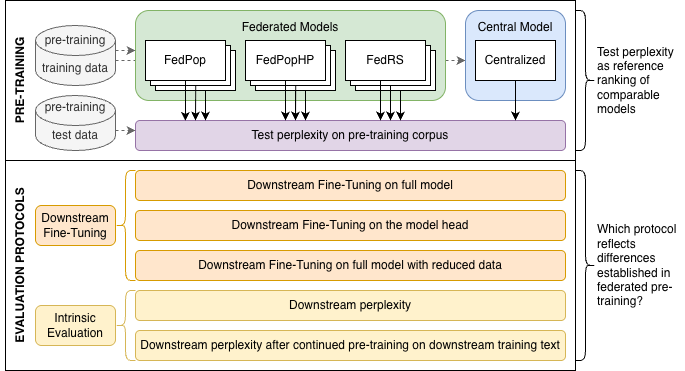}
  \caption{Overview of the considered evaluation protocols to measure federated pre-training quality.}
  \label{fig:overview}
\end{figure}

In our evaluation, we consider downstream fine-tuning using the GLUE \cite{GLUE} benchmark as well as next-token prediction on the GLUE text, as an intrinsic evaluation that is linked to the pre-training objective. Figure \ref{fig:overview} provides an overview of our workflow. Our results show that downstream fine-tuning can lead to misleading conclusions about pre-training quality, as it does not reliably preserve the reference ranking of the pre-trained models. By contrast, zero-shot next-token prediction on benchmark text better reflects this ordering.  These findings suggest that evaluation protocols closer to the pre-training objective can provide a more faithful signal for comparing federated pre-training strategies.

\section{Related Work}
Utilizing federated learning and Foundation models has been addresses in various ways \cite{ren2025advances,li2024synergizing}, but federated pre-training from scratch remains less explored.
Existing work shows feasibility across cross-device and cross-silo  scenarios, including decoder-only pre-training from scratch \cite{charlesFederatedFoundationModels}, pre-training a BERT and GPT-2 model using federated and split learning \cite{tianFedBERTWhenFederated2022}, studies of model size, heterogeneity and participation \cite{saniFutureLargeLanguage2024a}, large-scale federated pre-training \cite{saniPhotonFederatedLLM2024} and efficient transformer model pre-training using a federated tuning-while-training method \cite{grosser2025study}.
These works focus mainly on training feasibility, personalization, system design, scale or optimization. Evaluation is typically aligned with the respective modeling setup: Within the aboved mentioned works, \cite{charlesFederatedFoundationModels} uses both pre- and post-personalization, including task-specific personalization, and reports the corresponding loss, while \cite{tianFedBERTWhenFederated2022} evaluates BERT-style models via downstream fine-tuning and GPT-style models via perplexity. The remaining works \cite{saniFutureLargeLanguage2024a,saniPhotonFederatedLLM2024,grosser2025study} focus solely on perplexity-based evaluation. These differences highlight that evaluating federated pre-trained models is not uniform across the literature and that different protocols may capture different aspects of model quality.
In this work, we therefore compare downstream fine-tuning-based evaluation with next-token prediction as an intrinsic evaluation signal that remains closer to the pre-training objective.

\section{Experimental Setup} \label{sec:experimental_setup}
We compare a set of downscaled GPT-2 transformer models, which were pre-trained in a centralized and federated scenario. We assess different evaluation protocols by how they preserve differences of the test perplexity established after pre-training on its test set and report the Pearson and Spearman correlation between the evaluation protocol and the test perplexity after pre-training. We consider variants of downstream fine-tuning on the GLUE benchmark and variants of intrinsic evaluation, i.e. next-token prediction using the GLUE test text.

\subsection{Pre-training Baseline} \label{sec: pre-train baseline}
The models are given by the setup as described in \cite{grosser2025study}: The architecture is a downscaled version of GPT-2 \cite{GPT2} with a token length of 128, an embedding dimension of 256 as well as four layers and heads, resulting in a 16M parameter model. Pre-training was done using a subset of eight categories (as in \cite{saniFutureLargeLanguage2024a,saniPhotonFederatedLLM2024}) of the ThePile \cite{ThePile} dataset, such that each client holds 1024 training, 128 validation and 128 test sentences of at most 128 tokens per sentence. For tokenization, the GPT-2 tokenizer was used. Per category, 100 clients are created, giving a training set of 105M tokens. We run our experiments on NVIDIA A5000 GPUs.

The considered federated-trained models can be categorized in three main approaches, where each represents a different federated hyperparameter tuning approach. The difference between these approaches only lies at the hyperparameter selection strategy. Seven models were trained using FedPop \cite{FedPop} as the hyperparameter tuning strategy, while twelve models (FedPopHP) were trained with different parameterizations of FedPop itself. In addition, seven models (FedRS) were tuned using a simple random search approach.
All approaches were trained with a total communication budget of 3000 rounds, using the same hyperparameter search space and identical local hyperparameters: learning rate, the second momentum parameter of AdamW, gradient accumulation steps, and the number of local epochs. Due to the same training configuration and  procedure, the federated-trained models and their evaluation using the test dataset of all clients provide a fair comparison baseline. Additionally, one model is trained centrally on the same pre-training data corpus.

We define \emph{test perplexity} as the evaluation result on the test split of the considered ThePile subvariant obtained after pre-training. This metric serves as a controlled reference point for comparing models before any further adaptation. The resulting post-pre-training performance establishes an initial ranking of the models, and a reliable evaluation protocol should preserve a similar ordering.

To measure the impact of the model size, we consider four variants of downscaled GPT2 models: a 16M model with the same architecture as the federated-trained models, as well as a 32M, 64M and 124M parameter variant. To mimic pre-training of the first original GPT models \cite{GPT1,GPT2}, we pre-train our models on the BookCorpus2 dataset, available within the ThePile dataset. These models are not pre-trained or evaluated on the considered subvariant of the ThePile as the other models, therefore the \emph{test perplexity} is not provided.

\subsection{Downstream Fine-tuning}
We use the GLUE benchmark \cite{GLUE} to evaluate the downstream performance of the considered models. We exclude the WNLI task due to an adversarial split in the train, validation and test set \cite{wang2019superglue}.
We evaluate downstream performance under different fine-tuning regimes to examine how strongly the quality of pre-trained representations transfers to downstream tasks. In particular, we consider settings that vary the degree of task-specific adaptation.

We study full fine-tuning, where all model parameters are updated on the downstream task. This setting reflects the transfer learning scenario, in which pre-training provides the initialization for the full model. We further consider head-only fine-tuning, where the pre-trained backbone is frozen and only the task-specific head is trained. This variant allows us to assess utility of the learned representations more directly, as downstream performance depends primarily on the fixed pre-trained features.

In addition, we evaluate full fine-tuning with reduced downstream training data. To this end, we subsample the original GLUE training set and retain only 10\%, 1\%, and 0.1\% of the data, corresponding to reductions of 90\%, 99\%, and 99.9\%, respectively. This setting increases the dependence on pre-trained representations by limiting the amount of task-specific supervision available during fine-tuning.
For the data reduction, only the training dataset is reduced using stratified sampling for classification and binning for regression tasks.

Since the GLUE benchmark consists of classification and regression tasks, the model heads are adapted and fine-tuned using the same hyperparameters as in \cite{GPT1}. The maximum input length is reduced to 128 tokens to adjust with the architecture of the considered models. We report the overall GLUE score\footnote{Since we do not consider the WNLI task, the results for the WNLI testset was always set to the same result for all models to calculate the GLUE score.} calculated on the test sets of the benchmark, and all individual results are given in the appendix.

\subsection{Intrinsic Evaluation Signals}
As downstream evaluation requires supervised adaptation to certain tasks, we additionally consider an evaluation protocol that stays closer to the original pre-training objective. As intrinsic evaluation of the considered pre-trained language models we measure the next-token prediction of our models by calculating the perplexity for each GLUE task using their test text\footnote{CoLA contains intentionally ungrammatical sentences, and QQP includes informal or misspelled forum text. We acknowledge that next-token prediction on these tasks should be considered with caution.}. This enables a direct evaluation of the model without introducing a task-specific adaptation and reflects the ability to perform next-token prediction on an held-out benchmark text, which was not seen during pre-training.
We further consider two settings of this intrinsic evaluation: a zero-shot setting, where we directly apply next-token prediction using the models as directly provided after pre-training, as well as continuing pre-training on the training text for each GLUE task before evaluating the next-token prediction.

For both settings, we report the \emph{average perplexity} over the individual perplexities across GLUE tasks.

\section{Results}
We provide the results of our experiments as described in section \ref{sec:experimental_setup}.
\subsection{Fine-tuning on Downstream tasks}
Table \ref{tab:scores_full_head-only} reports the GLUE scores for full fine-tuning and head-only fine-tuning along with the test perplexities from pre-training for our set of federated-trained models, and with a centralized model. The \emph{test perplexities} range between $215.59$ and $384.42$ for the federated-trained models and $89.07$ for the centralized model. In contrast, downstream performance varies only within a narrow band: GLUE scores range from $50.0$ to $53.1$ under full fine-tuning and from $37.7$ to $41.4$ under head-only fine-tuning.
Thus, while the pre-training \emph{test perplexity} differentiates clearly between pre-training models, these differences are not preserved by downstream fine-tuning. Models that are well separated in test perplexity achieve very similar downstream scores, and the resulting ordering does not match the pre-training ranking.

\begin{table}[h]
    \centering
\begin{tabular}{lccc}
\toprule
\textbf{Method} & \textbf{Test Ppl} & \multicolumn{2}{c}{\textbf{GLUE Scores}} \\
\cmidrule(lr){3-4}
 & & Full FT & Head Only FT\\
\midrule
Centralized & 89.07 & 51.9 & 39.1 \\
\midrule
FedPopHP & 215.59 & 52.3 & 39.7 \\
 & 221.34 & 52.9 & 39.7 \\
  & 223.14 & 52.6 & 40.6 \\
  & 223.96 & 52.4 & 40.2 \\
  & 224.21 & 51.9 & 40.8 \\
  & 231.66 & 53 & 39.8 \\
  & 235.06 & 52.6 & 38.4 \\
  & 236.57 & 51.5 & 39.8 \\
  & 238.35 & 51.6 & 39 \\
  & 238.56 & 52.3 & 38.3 \\
  & 255.59 & 52.3 & 37.7 \\
  & 260.36 & 52.7 & 38.1 \\
\midrule
FedPop & 238.37 & 53.1 & 40.6 \\
   & 240.46 & 52.5 & 40.2 \\
   & 247.24 & 52.6 & 40.3 \\
   & 252.29 & 53 & 38.2 \\
  & 260.85 & 52.1 & 40.3 \\
   & 268.63 & 52.3 & 39.8 \\
   & 269.99 & 52 & 40.1 \\
\midrule
FedRS & 238.81 & 53 & 40 \\
 & 287.53 & 52.4 & 40.4 \\
 & 309.16 & 50 & 40.4 \\
   & 325.47 & 52.3 & 41.4 \\
   & 342.56 & 50.7 & 40.5 \\
   & 355.85 & 51.7 & 41.2 \\
   & 384.42 & 52.5 & 40.8 \\
\bottomrule
\end{tabular}
    \caption{\emph{Test perplexity} after pre-training (Test ppl) and GLUE scores of Full Fine-Tuning and Head-Only Fine-Tuning for a 16M transformer model on federated-trained models (FedPopHP, FedPop, FedRS) and a central trained model.}
    \label{tab:scores_full_head-only}
\end{table}

To further examine the role of downstream adaption, we additionally consider GPT-style models of smaller sizes as described in \ref{sec: pre-train baseline}.  We provide detailed results for all tasks in the appendix in Table \ref{tab:appendix_finetuning}. For full fine-tuning on these four models, the GLUE scores range between $51.6$ and $52.6$, with no consistent order on the model size. We further evaluate five randomly initialized 16M models under full fine-tuning, obtaining GLUE scores between $50.9$ and $52.6$. Thus, all considered alternative models achieve similar results to the federated 16M models.

We reduce the amount of downstream training examples to test whether the masking effect of fine-tuning persists in a lower-data regime.
Instead of using the whole GLUE data corpus for fine-tuning, we reduce the size of the fine-tuning dataset by 90\%, 99\% and 99.9\% of its original size. We perform full fine-tuning with the reduced GLUE data corpora for the centrally trained model and for a subset of the federated-trained models, which are selected to cover a whole range of test perplexities. We report the results along with the pre-training \emph{test perplexities} in Table \ref{tab:scores_reduced_fine-tuning}.
With a 90\% reduction, GLUE scores still vary only in a narrow range between $45.3$ and $47.4$. With 99\% and 99.9\% reduction, the score range becomes wider, for the first from $33.9$ to $44.7$ and for the latter from $30.5$ to $37.3$, indicating that reduced supervision increases sensitivity to model differences. However, even in this strongly reduced setting, the downstream ranking does not recover the ordering established by pre-training \emph{test perplexity}.

Table \ref{tab:correlation} shows the pearson and spearman correlation for the evaluation protocols. The pearson and spearman correlation for full fine-tuning is $-0.28$ and $-0.25$, while for head-only fine-tuning the pearson correlation is $0.42$ and the spearman correlation is $0.38$. Head-only fine-tuning shows a moderate linear correlation, whereas full fine-tuning holds a smaller correlation in magnitude than head-only fine-tuning and is slightly negative. For fine-tuning on the reduced datasets, we observe no correlation (pearson $0.02$ and spearman $0.00$ for the 99.9\% data reduction) or a negative tendency (pearson $-0.56$ and spearman $-0.21$ for the 90\% data reduction). Note that for the data reduction correlation only 5 model are evaluated.

Overall, in the considered setting, fine-tuned downstream performance fails to preserve the differences observed after pre-training. Head-only fine tuning yields the highest correlation with the pre-training \emph{test perplexity} among the considered downstream evaluation protocols.

\begin{table}
    \centering
\begin{tabular}{lcccc}
\toprule
\textbf{Method} & \textbf{Test Ppl} & \multicolumn{3}{c}{\textbf{GLUE Scores reduced data}} \\
\cmidrule(lr){3-5}
 & & 90\% & 99\% & 99.9\% \\
\midrule
Central & 89.07 & 46.5 & 44.7 & 37.3 \\
\midrule
FedPopHP & 215.59 & 46.9 & 33.9 & 30.5 \\
\midrule
FedPop & 238.37 & 46.5 & 40.9 & 34.3 \\
\midrule
FedRS & 238.81 & 47.4 & 36.6 & 35.6 \\
 & 384.42 & 45.3 & 40 & 36.8 \\
\bottomrule
\end{tabular}
    \caption{\emph{Test perplexity} after pre-training (Test ppl) and GLUE scores of fine-tuning the full model on a strongly reduced GLUE data corpora for a subset of pre-trained models.}
    \label{tab:scores_reduced_fine-tuning}
\end{table}

\subsection{Intrinsic Evaluation Signals}
\begin{figure}
  \centering
  \includegraphics[width=0.9\linewidth]{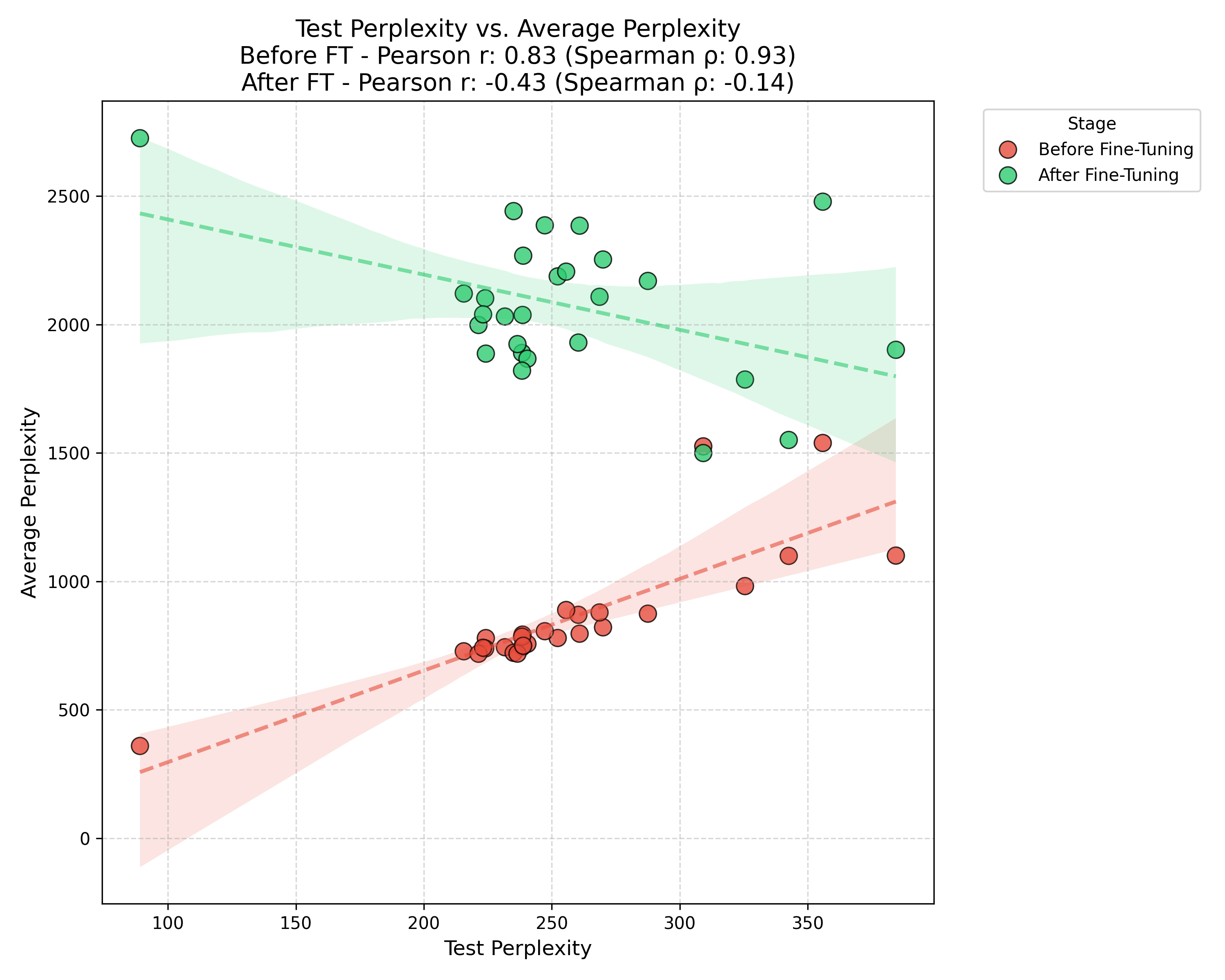}
  \caption{Relationship between the pre-training \emph{test perplexity} and average perplexity on GLUE data. Before fine-tuning (red) denotes direct next-token prediction on the pre-trained models showing a strong positive correlation with the \emph{test perplexity}. After fine-tuning (green) denotes continued pre-training on the GLUE training sets showing no clear correlation. Dashed lines give linear regression fits; shaded areas show bootstrapped 95\% confidence intervals.}
  \label{fig:correlation}
\end{figure}

As an alternative, we analyze intrinsic evaluation signals. Specifially, we use next-token prediction and report perplexity as it matches the pre-training object. We consider two variants: average perplexity directly on the raw models, and average perplexity after continuing the pre-training on the individual GLUE datasets. Figure \ref{fig:correlation} shows both quantities with the original pre-training \emph{test perplexity}. For reference, we add linear regression lines and bootstrapped 95\% confidence intervals. A table of the results can be found in the appendix in Table \ref{tab:appendix_perplexities}.

The results of the direct next-token prediction approach show a strong positive correlation between the \emph{test perplexity} and average perplexity on the GLUE dataset. As shown in Figure \ref{fig:correlation}, the models are tightly clustered around the regression line, indicating a strong linear relationship between \emph{test perplexity} and average perplexity on the GLUE texts. This observation is supported by a high Pearson correlation value of $r = 0.83$ and Spearman correlation value of $\rho = 0.93$, as seen in Table~\ref{tab:correlation}.

For continued pre-training, we observe a higher average perplexity after continued pre-training using the GLUE training text, including our GPT variants.
In contrast to the direct evaluation, this variant shows no clear relationship with the original \emph{test perplexity}. In Figure \ref{fig:correlation}, the models are widely spread around the regression line.
The corresponding correlation coefficients differ in magnitude (pearson -0.43 and spearman -0.14) with both coefficients being negative.
Other than next-token prediction in a zero-shot setting, the ranking induced by pre-training \emph{test perplexity} is not preserved after continued pre-training on GLUE for our models.

Although the compared models differ substantially in pre-training \emph{test perplexity}, these differences largely disappear after downstream adaptation, especially for full model fine-tuning. This indicates that supervised fine-tuning can overwrite or compensate for differences induced during pre-training.
The considered intrinsic evaluation on continued pre-training again indicates adaptation to a benchmark test text changes relative ordering of the models. In our experiments, intrinsic evaluation in form of direct next-token prediction on benchmark text preserves the reference ranking substantially better than downstream performance, even for head-only fine-tuning.
This suggests that evaluation protocols of zero-shot manner provide a more faithful signal of pre-training quality.

\begin{table}
\centering
\begin{tabular}{lcc}
\toprule
\textbf{Evaluation Protocol} & \textbf{Pearson} & \textbf{Spearman} \\
\midrule
GLUE Score Full FT & -0.28  & -0.25  \\
GLUE Score Head Only FT & 0.42  & 0.38  \\
\midrule
GLUE Score red. data 90\% & -0.56  & -0.21  \\
GLUE Score red. data 99\% & -0.34  & -0.30 \\
GLUE Score red. data 99.9\% & 0.02  & 0.00 \\
\midrule
GLUE Avg Ppl Direct  & 0.83  & 0.93  \\
GLUE Avg Ppl Cont.  & -0.43 & -0.14 \\
\bottomrule
\end{tabular}
\caption{Pearson and Spearman correlation between the \emph{test perplexity} after pre-training and the evaluation protocols: Downstream performance on the GLUE benchmark with full and head-only fine-tuning (FT), including full fine-tuning on reduced training data as well as intrinsic evaluation using the average perplexity on next-token prediction of the GLUE test corpora, measured directly and after continued (cont.) pre-training on the GLUE train corpora.}
\label{tab:correlation}
\end{table}

\section{Limitations}
Our findings are based on controlled experiments at relatively small scale. In particular, we study a 16M parameter transformer model trained in centralized and federated settings on a small subset of ThePile. While this setup enables a controlled comparison across models, it is substantially smaller than modern foundation model pre-training regimes. We further restrict evaluation to the GLUE benchmark and to next-token prediction on the corresponding benchmark text, which covers only a limited range of downstream and intrinsic evaluation settings.

In addition, our analysis measures evaluation quality by how well a protocol preserves a reference ranking induced by the \emph{test perplexity} after pre-training. This is a useful criterion in our controlled setting, but it does not capture all possible notions of model quality. Different conclusions may arise for larger models, different benchmark families, broader domains, or alternative definitions of pre-training quality. Extending the analysis to such settings remains important future work. In particular, it remains important future work to examine whether our findings hold for substantially larger language models. Moreover, our study is limited to the language domain and does not address whether similar patterns arise in other modalities, such as images or time series.

\section{Conclusion}
We studied how to evaluate federated pre-training quality under controlled conditions. Using a set of federatedly pre-trained transformer models that are directly comparable, we assessed evaluation protocols by whether they preserve a reference ranking established on the pre-training testset. Our results show that downstream fine-tuning on GLUE, including head-only and reduced-data variants, does not reliably preserve this ranking. We observed similar results for next-token prediction on models when pre-training was continued on a held-out benchmark text. In contrast, direct next-token prediction on benchmark text without further training shows a strong correspondence with the pre-training ordering of the models.

These findings suggest that the choice of evaluation protocol matters substantially when comparing federated pre-training methods. While downstream fine-tuning remains useful for measuring end-task performance, it may be insufficient as the only basis for assessing pre-training quality, since it can confound the effects of pre-training with those of task-specific adaptation. If the goal is to evaluate the pre-training procedure itself, protocols that minimize or avoid additional training may provide a more faithful comparison.

\appendix

\section{Full experiment results}
We provide more detailed results of our experiments. Table \ref{tab:appendix_finetuning} shows the GLUE results for full fine-tuning and head-only fine-tuning. To save space, we show the results from head-only fine-tuning in an extra row in brackets. Table \ref{tab:appendix_reduction} shows the details of the GLUE results for the considered reduced data setting. Table \ref{tab:appendix_perplexities} shows the average perplexities over the GLUE test text when directly calculated on the models and after continued pre-training on the GLUE train text.

\begin{table*}
    \centering
\scriptsize
\begin{tabular}{lcccccccccc}
\toprule
\textbf{Method} & \textbf{Test PPL} & \textbf{CoLA} & \textbf{SST-2} & \textbf{MRPC} & \textbf{STS-B} & \textbf{QQP} & \textbf{MNLI (m/mm)} & \textbf{QNLI} & \textbf{RTE} & \textbf{GLUE Score} \\
\midrule
FedPop & 238.37 & 12 & 82.4 & 80.3 & 32.9 & 55.5 & 61/60.9 & 61.2 & 50 & 53.1 \\
 & & (0) & (54.6) & (78.8) & (2.5) & (30.7) & (36.9/38) & (56.1) & (49.5) & (40.6) \\
  & 240.46 & 8.1 & 83.1 & 79.9 & 30.8 & 54.8 & 61.5/61.5 & 59.5 & 52 & 52.5 \\
 & & (3) & (55.6) & (74.7) & (1.1) & (24.2) & (37.1/37.6) & (55.5) & (50.5) & (40.2) \\
  & 247.24 & 5.9 & 82.3 & 78.3 & 34.2 & 54.6 & 60.9/60.4 & 61.8 & 51.8 & 52.6 \\
 & & (4.2) & (54.8) & (76.5) & (7.5) & (10.8) & (36.7/38) & (54.6) & (49.4) & (40.3) \\
  & 252.29 & 7 & 82.7 & 80 & 34.4 & 56 & 60.9/60.6 & 61.1 & 51.1 & 53 \\
 & & (1.3) & (52.1) & (70.9) & (-2) & (13.2) & (47.4/37.7) & (54.8) & (50) & (38.2) \\
  & 260.85 & 8.6 & 81.7 & 80.5 & 29.1 & 53.2 & 60.7/59.9 & 60.9 & 51.9 & 52.1 \\
 & & (0.6) & (55.5) & (75.5) & (4.1) & (27.9) & (37.2/36.6) & (54.6) & (50.1) & (40.3) \\
  & 268.63 & 7.1 & 80.7 & 80.9 & 31.7 & 54.4 & 60.8/60.7 & 61.4 & 50.7 & 52.3 \\
 & & (0) & (52.7) & (78.2) & (1.8) & (19.5) & (37.5/38.2) & (55.5) & (49.8) & (39.8) \\
  & 269.99 & 4.7 & 82.2 & 77.5 & 32.9 & 54.6 & 61.1/60.4 & 61.2 & 50.4 & 52 \\
 & & (2.1) & (54.4) & (75.9) & (6.3) & (19) & (36.7/37.8) & (54.5) & (48.7) & (40.1) \\
\midrule
FedPopHP & 215.59 & 7.6 & 82.8 & 76.6 & 32 & 55.5 & 61.6/60.9 & 61 & 51.2 & 52.3 \\
 & & (-1.7) & (54.4) & (72.4) & (1.2) & (27.8) & (37.3/37.6) & (56.3) & (50.3) & (39.7) \\
  & 221.34 & 9.4 & 82 & 78.4 & 34.2 & 54.9 & 61.3/60.8 & 60.8 & 52.4 & 52.9 \\
 & & (3.2) & (54.3) & (74.8) & (-3.5) & (25.3) & (37.5/37.8) & (56) & (50.2) & (39.7) \\
  & 223.14 & 8.3 & 82.5 & 77.9 & 32.4 & 57.3 & 61.5/60.9 & 60.4 & 50.5 & 52.6 \\
 & & (4.3) & (54.8) & (72.7) & (4.2) & (28.6) & (37.2/38.7) & (56.8) & (48) & (40.6) \\
  & 223.96 & 8.4 & 82.2 & 77 & 31.8 & 55.2 & 61.3/60.9 & 61 & 50.8 & 52.4 \\
 & & (-0.7) & (55) & (76.1) & (2.6) & (27.1) & (37.7/39.2) & (56) & (49.9) & (40.2) \\
  & 224.21 & 4.7 & 82.4 & 73 & 32.8 & 57.1 & 61.7/61 & 61.7 & 50.4 & 51.9 \\
 & & (0) & (53.9) & (78.4) & (8.1) & (24.4) & (37.6/37.3) & (56.5) & (49.9) & (40.8) \\
  & 231.66 & 9.4 & 82 & 80.7 & 30.7 & 55.3 & 61.5/60.9 & 61.4 & 52.9 & 53 \\
 & & (-4.1) & (54.3) & (68.3) & (10.9) & (25.4) & (37.9/37.6) & (56.5) & (49.3) & (39.8) \\
  & 235.06 & 9.6 & 83 & 78.4 & 30.4 & 54.8 & 61.4/61.2 & 61 & 51.1 & 52.6 \\
 & & (-1.4) & (54.1) & (56.9) & (-0.4) & (23.4) & (37.7/38.2) & (55.4) & (50.3) & (38.4) \\
  & 236.57 & 5 & 81.7 & 72.7 & 32.4 & 54 & 61.6/61.4 & 60.9 & 50.6 & 51.5 \\
 & & (0.4) & (55.6) & (70.2) & (-0.1) & (32.4) & (37.8/38.1) & (55.9) & (50.5) & (39.8) \\
  & 238.35 & 6.1 & 82 & 71 & 32.9 & 55.5 & 60.4/60.2 & 60.6 & 50.7 & 51.6 \\
 & & (-1.2) & (54) & (63) & (10.9) & (21.2) & (37/37) & (55.4) & (47.5) & (39) \\
  & 238.56 & 6.9 & 82.1 & 80 & 29.9 & 54.9 & 61.5/61 & 61.4 & 50.7 & 52.3 \\
 & & (-4.8) & (54.8) & (76.6) & (-6.9) & (20.9) & (37.5/38) & (54.7) & (49.6) & (38.3) \\
  & 255.59 & 12.3 & 81.4 & 80.5 & 29.5 & 53.2 & 60.5/59.4 & 61.5 & 50.2 & 52.3 \\
 & & (-1.3) & (53.6) & (59.5) & (-1) & (21.2) & (36.9/37.8) & (55.5) & (49.6) & (37.7) \\
  & 260.36 & 13.9 & 81.6 & 76.7 & 31.7 & 53.9 & 61.4/60.4 & 61.1 & 51.2 & 52.7 \\
 & & (-0.6) & (55.3) & (72.7) & (-8.7) & (13.5) & (37/37.6) & (56.3) & (50.8) & (38.1) \\
\midrule
FedRS & 238.81 & 11.2 & 81.6 & 78.5 & 34 & 54.7 & 61.4/60.5 & 61 & 52.3 & 53 \\
 & & (1.9) & (53.9) & (71.7) & (6.2) & (17.9) & (37.3/38.3) & (55.4) & (52.2) & (40) \\
  & 287.53 & 9.4 & 81.5 & 77.5 & 33.1 & 54.1 & 60.8/60.7 & 61.4 & 50.7 & 52.4 \\
 & & (3) & (54.3) & (74.3) & (10.3) & (15.2) & (37/37.3) & (55.1) & (49.6) & (40.4) \\
  & 309.16 & 0 & 80.7 & 79.7 & 20.8 & 54.3 & 57.9/57.6 & 60.4 & 52 & 50 \\
 & & (0) & (52.7) & (79.4) & (-0.3) & (33.9) & (36.9/37.4) & (56.9) & (51.3) & (40.4) \\
  & 325.47 & 7.8 & 83 & 72.9 & 32.6 & 55.9 & 63.5/63.5 & 60.4 & 50.8 & 52.3 \\
 & & (-1.9) & (54.6) & (73.2) & (19.2) & (23.3) & (37.1/38.3) & (56.9) & (48.6) & (41.4) \\
  & 342.56 & 0 & 81.1 & 78.8 & 26.2 & 56.2 & 57.6/57.4 & 61.2 & 50.3 & 50.7 \\
 & & (0) & (55.9) & (78.5) & (5.3) & (25.6) & (36.2/36.6) & (55.9) & (49.2) & (40.5) \\
  & 355.85 & 7.3 & 80.2 & 80.3 & 30 & 53.6 & 60.4/59.4 & 60.4 & 50.5 & 51.7 \\
 & & (0) & (53.9) & (79.6) & (14) & (24.7) & (34.9/34.6) & (55.7) & (49.9) & (41.2) \\
  & 384.42 & 7.9 & 83.5 & 76.7 & 31.8 & 55.8 & 62/62.6 & 61 & 50.2 & 52.5 \\
 & & (-1.9) & (55.2) & (76.2) & (7) & (24.6) & (37.3/37.8) & (56.7) & (51.3) & (40.8) \\
\midrule
Centralized & 89.07 & 6.1 & 81.4 & 77.5 & 30.2 & 54.4 & 62.1/61.4 & 60.1 & 51.5 & 51.9 \\
 & & (-0.3) & (55.7) & (65.6) & (-2.5) & (33.4) & (36.6/36.9) & (56.1) & (48) & (39.1) \\
\midrule
Random Init & --- & 4.8 & 81.5 & 69.2 & 30.1 & 53 & 60.4/59.8 & 60.5 & 52.4 & 50.9 \\
   & --- & 11.1 & 81 & 76 & 31 & 55.9 & 60/60.2 & 60.4 & 52.1 & 52.6 \\
   & --- & 8.2 & 81.3 & 71.5 & 30.3 & 54.5 & 60/59.6 & 59.8 & 51.7 & 51.6 \\
   & --- & 11.1 & 82.3 & 74.5 & 28.8 & 55.2 & 60.4/60.4 & 60.2 & 52.2 & 52.5 \\
   & --- & 9.4 & 81.9 & 77.7 & 28.8 & 55.1 & 60.5/60.4 & 60.4 & 51 & 52.3 \\
\midrule
GPT 16M & --- & 6.2 & 80.3 & 75.1 & 33.8 & 56.5 & 60.5/60.8 & 60.0 & 52.2 & 52.2 \\
GPT 32M & --- & 6.9 & 80.4 & 74.0 & 27.6 & 56.9 & 61.9/62.0 & 60.0 & 51.9 & 51.6 \\
GPT 64M & --- & 10.8 & 80.5 & 76.9 & 26.5 & 58.5 & 61.7/61.4 & 58.8 & 52.2 & 52.3 \\
GPT 124M & --- & 10.2 & 80.5 & 76.7 & 28.5 & 58.5 & 60.7/60.9 & 59.8 & 53.2 & 52.6 \\
\bottomrule
\end{tabular}
    \caption{Detailed results on the GLUE benchmark for full fine-tuning and head-only fine-tuning (reported in brackets).}
    \label{tab:appendix_finetuning}
\end{table*}

\begin{table*}
\centering
\tiny
\begin{tabular}{lcccccccccc}
\toprule
\textbf{Method} & \textbf{Test PPL} & \textbf{CoLA} & \textbf{SST-2} & \textbf{MRPC} & \textbf{STS-B} & \textbf{QQP} & \textbf{MNLI (m/mm)} & \textbf{QNLI} & \textbf{RTE} & \textbf{GLUE Score} \\
\midrule
\multicolumn{11}{c}{\textbf{90\% Reduction}} \\
\midrule
FedPop & 238.37 & -3 & 72.4 & 78.2 & 14 & 49.6 & 52/51.3 & 58.9 & 49.3 & 46.5 \\
\midrule
FedPopHP & 215.59 & -0.8 & 72.1 & 80.1 & 15.2 & 49.2 & 51.9/51.2 & 58.7 & 49.4 & 46.9 \\
\midrule
FedRS & 238.81 & 5.2 & 72.9 & 69.8 & 20 & 49.5 & 52.4/53.2 & 60 & 50.2 & 47.4 \\
 & 384.42 & 0 & 63.6 & 76.7 & 12.7 & 51.1 & 46.4/48.8 & 58.6 & 49.7 & 45.3 \\
\midrule
Centralized & 89.07 & 5.2 & 74.8 & 72.6 & 13.5 & 49 & 50.2/50 & 56.9 & 49 & 46.5 \\
\midrule
\multicolumn{11}{c}{\textbf{99\% Reduction}} \\
\midrule
FedPop & 238.37 & 0.6 & 52.8 & 67 & 4.6 & 42.2 & 40.2/42.8 & 57 & 50.9 & 40.9 \\
\midrule
FedPopHP & 215.59 & 0.4 & 50.8 & 11.1 & -8.5 & 42.8 & 38.8/41.2 & 54.4 & 50 & 33.9 \\
\midrule
FedRS & 238.81 & 1.6 & 55 & 35.3 & -13 & 40.9 & 41.4/43.2 & 55.2 & 51 & 36.6 \\
 & 384.42 & 0 & 51.4 & 76.4 & -1 & 37 & 36.7/37.2 & 55.9 & 49.6 & 40 \\
\midrule
Centralized & 89.07 & 0.7 & 50.7 & 70.9 & -6.5 & 43.4 & 39/40 & 53.3 & 50.3 & 44.7 \\
\midrule
\multicolumn{11}{c}{\textbf{99.9\% Reduction}} \\
\midrule
FedPop & 238.37 & -0.8 & 49.8 & 7.9 & -1 & 32.9 & 33.6/33.6 & 51.3 & 50 & 34.3 \\
\midrule
FedPopHP & 215.59 & -3.9 & 49.9 & 0 & 3.9 & 34.1 & 32.3/31.8 & 51.1 & 47.9 & 30.5 \\
\midrule
FedRS & 238.81 & -6.6 & 51.2 & 76.8 & -6.2 & 34.4 & 32.1/31.1 & 51.6 & 50 & 35.6 \\
 & 384.42 & 0.3 & 49 & 61.5 & 1.2 & 15.6 & 33.3/34.5 & 53.2 & 50.2 & 36.8 \\
\midrule
Centralized & 89.07 & -4.7 & 50.5 & 79.4 & NaN & 7.4 & 33.4/33.3 & 50.8 & 50.1 & 37.3 \\
\bottomrule
\end{tabular}
\caption{Detailed results on the GLUE benchmark for full fine-tuning with reduced data.}
\label{tab:appendix_reduction}
\end{table*}

\begin{table*}
\centering
\tiny
\begin{tabular}{lccc|ccc|ccc|ccc}
\toprule
\multicolumn{4}{c}{\textbf{Centralized}} & \multicolumn{3}{c}{\textbf{FedPopHP}} & \multicolumn{3}{c}{\textbf{FedPop}} & \multicolumn{3}{c}{\textbf{FedRS}} \\
\cmidrule(lr){1-4} \cmidrule(lr){5-7} \cmidrule(lr){8-10} \cmidrule(lr){11-13}
\textbf{Model} & \textbf{Test} & \textbf{Direct} & \textbf{Cont.} & \textbf{Test} & \textbf{Direct} & \textbf{Cont.} & \textbf{Test} & \textbf{Direct} & \textbf{Cont.} & \textbf{Test} & \textbf{Direct} & \textbf{Cont.} \\
\midrule
Centralized & 89.07 & 359.04 & 2725.09 & 215.59 & 727.61 & 2120.43 & 238.37 & 743.92 & 1889.18 & 238.81 & 749.23 & 2267.73 \\
GPT 16M & --- & 901.48 & 2055.49 & 221.34 & 718.01 & 1997.96 & 240.46 & 756.47 & 1866.64 & 287.53 & 874.06 & 2169.45 \\
GPT 32M & --- & 1220.71 & 4605.21 & 223.14 & 741.27 & 2039.35 & 247.24 & 805.80 & 2386.02 & 309.16 & 1525.41 & 1499.44 \\
GPT 64M & --- & 2386.59 & 2115.93 & 223.96 & 738.63 & 2102.08 & 252.29 & 779.35 & 2187.27 & 325.47 & 981.46 & 1785.65 \\
GPT 124M & --- & 807.58 & 3013.41 & 224.21 & 779.01 & 1886.47 & 260.85 & 796.10 & 2384.58 & 342.56 & 1099.03 & 1550.33 \\
    &     &     &     & 235.06 & 721.67 & 2441.27 & 269.99 & 820.88 & 2253.00 & 384.42 & 1100.15 & 1901.32 \\
    &     &     &     & 236.57 & 718.43 & 1923.53 &     &     &     &     &     &     \\
    &     &     &     & 238.35 & 784.57 & 1820.10 &     &     &     &     &     &     \\
    &     &     &     & 238.56 & 792.62 & 2037.08 &     &     &     &     &     &     \\
    &     &     &     & 255.59 & 888.34 & 2205.64 &     &     &     &     &     &     \\
    &     &     &     & 260.36 & 869.67 & 1929.56 &     &     &     &     &     &     \\
\bottomrule
\end{tabular}
\caption{Detailed results on the average GLUE perplexities when directly calculated on the model and after continued pre-training on the GLUE datasets. Test refers to the pre-training test perplexity. Centralized trained models are added for reference.}
\label{tab:appendix_perplexities}
\end{table*}

\section*{Acknowledgments}
This research was supported by the German Federal Ministry for Economic Affairs and Energy under grant 01MD23001C (OpenFLaaS) at Siemens AG.

\section*{Contribution Statement}
We mark equal contributions of authors with a \textbf{*} symbol.

\bibliographystyle{named}
\bibliography{ijcai26}

\typeout{get arXiv to do 4 passes: Label(s) may have changed. Rerun}

\end{document}